\documentclass[journal]{IEEEtran}
\usepackage{graphicx}
\usepackage{glossaries}
\usepackage{cite}
\usepackage{picinpar}
\usepackage{amsmath}
\usepackage{url}
\usepackage{booktabs}
\usepackage{flushend}
\usepackage[utf8]{inputenc}
\usepackage{colortbl}
\usepackage{soul}
\usepackage{stfloats}
\usepackage{multirow}
\usepackage{pifont}
\usepackage{color}
\usepackage{alltt}
\usepackage[hidelinks]{hyperref}
\usepackage{enumerate}
\usepackage{siunitx}
\usepackage{breakurl}
\usepackage{epstopdf}
\usepackage{amsmath}
\usepackage{amsfonts}
\usepackage{amsbsy}
\usepackage{amssymb}
\usepackage{pbox}
\usepackage{subcaption}
\usepackage{xcolor} 
\usepackage{algorithm}
\usepackage{algpseudocode}
\usepackage{fancyhdr}

\newtheorem{definition}{\bf Definition}

\newcount\Comments  
\definecolor{darkgreen}{rgb}{0,0.5,0}
\definecolor{purple}{rgb}{1,0,1}
\newcommand{\kibitz}[2]{\ifnum\Comments=1\textcolor{#1}{#2}\fi}

\fancypagestyle{firstpage}{
  \fancyhf{}  
  \fancyhead[C]{\textcolor{red}{\textbf{Preprint accepted by IEEE Transactions on Industrial Cyber-Physical Systems. To appear in TICPS on IEEE Explore.}}}
}

\begin{document}

\title{Hierarchical Testing with Rabbit Optimization for Industrial Cyber-Physical Systems}

\author{
	\vskip 1em
	
	Jinwei Hu$^{1\S}$, Zezhi Tang$^{2\S}$, Xin Jin$^2$, Benyuan Zhang$^2$, Yi Dong$^{1}$, Xiaowei Huang$^1$

	\thanks{
	
		
  
            \S \quad equal contributions


		1. Department of Computer Science, University of Liverpool, L69 7ZX, the UK (e-mail: jinwei.hu, yi.dong, xiaowei.huang@liverpool.ac.uk). 
		
		2. Department of Electrical and Electronic Engineering, University of Sheffield, S10 2TN, the UK(e-mail: zezhi.tang, xjin32, bzhang79@sheffield.ac.uk).
	}
}

\maketitle
\thispagestyle{firstpage}

\begin{abstract}
This paper presents HERO (Hierarchical Testing with Rabbit Optimization), a novel black-box adversarial testing framework for evaluating the robustness of deep learning-based Prognostics and Health Management systems in Industrial Cyber-Physical Systems. Leveraging Artificial Rabbit Optimization, HERO generates physically constrained adversarial examples that align with real-world data distributions via global and local perspective. Its generalizability ensures applicability across diverse ICPS scenarios. This study specifically focuses on the Proton Exchange Membrane Fuel Cell system, chosen for its highly dynamic operational conditions, complex degradation mechanisms, and increasing integration into ICPS as a sustainable and efficient energy solution. Experimental results highlight HERO's ability to uncover vulnerabilities in even state-of-the-art PHM models, underscoring the critical need for enhanced robustness in real-world applications. By addressing these challenges, HERO demonstrates its potential to advance more resilient PHM systems across a wide range of ICPS domains.
\end{abstract}


{}

\definecolor{limegreen}{rgb}{0.2, 0.8, 0.2}
\definecolor{forestgreen}{rgb}{0.13, 0.55, 0.13}
\definecolor{greenhtml}{rgb}{0.0, 0.5, 0.0}

\section{Introduction}

With the rapid development of net zero, there is a need for advanced predictive models and system integration plays a crucial role in the field of renewable energy technologies, particularly in the deployment and management of Proton Exchange Membrane Fuel Cells (PEMFC). Regarded as an integral part of future energy conversion technologies, PEMFC boast high energy conversion efficiency, low operating temperature, low emissions, and rapid startup capabilities \cite{barbir1997efficiency}. Their application range, from portable power sources to transportation and stationary power generation systems, displays their exceptional flexibility and environmental friendliness, indicating their pivotal role in future energy systems \cite{wee2007applications}. However, the widespread deployment of PEMFC across various sectors brings significant challenges in ensuring their long-term stable operation and accurately approximating their lifespan. This highlights the critical role of Prognostics and Health Management (PHM) strategies, which employ predictive models to forecast system degradation and estimate Remaining Useful Life (RUL), thereby enabling preventive maintenance in critical industrial applications \cite{sutharssan2017review}.



Machine Learning (ML) and Deep Learning (DL) models are always employed in enhancing the accuracy and robustness of detections in PHM systems, capitalizing on their ability to discern complex data patterns from extensive datasets \cite{fink2020potential}. Despite the significant advantages these models provide in refining predictions and operational efficiencies, they inherently exhibit vulnerabilities that can undermine their effectiveness in critical applications such as PEMFC lifespan predictions. One significant challenge is the ``black-box" nature of many deep learning algorithms. This complexity complicates the interpretability and validation of the models, raising concerns in critical applications like PEMFC operation monitoring \cite{ding2022guiding, ao2022diagnosis}. While ML and DL models always perform well under typical conditions in ICPS, they often lack robustness against diverse threats, such as cyber attacks \cite{wu2023sir} and adversarial attacks \cite{gao2017deepcloak}. For example, subtle, intentionally crafted disturbances in PHM system inputs can cause significant prediction errors for PEMFC health and performance, potentially leading to undetected system failures or suboptimal operations until severe issues arise \cite{9356218, basak2021universal}.

Adversarial testing, a technique widely applied across various fields, is focusing on identifying performance-degrading examples that can impact DL models. Identified adversarial examples can also facilitate system improvements through adaptation or retraining \cite{yaghoubi2019gray, dreossi2018counterexample}. Mainstream white-box testing methods, such as DLFuzz \cite{9099600} and DeepXplore \cite{pei2017deepxplore}, predominantly employ coverage-guided testing to actively mutate inputs, maximizing neuron coverage and exposing prediction discrepancies. However, these methods are primarily designed for the image domain and do not cater to the nuances of time-series DL models. Transitioning to time-series DL models, TESTRNN \cite{huang2021coverage} is also a coverage-guided testing approach that designs metrics to assess both values and temporal relationships in LSTM inputs. Despite its innovative approach, TESTRNN primarily focuses on image data and is not suitable for the tabular data typically associated with PHM systems. Moreover, because of its model-specific nature, coverage-guided testing does not adapt well to applications involving transformer-based DL tools. In black-box adversarial testing, the Hierarchical Distribution-Aware (HDA) \cite{huang2023hierarchical} approach employs a dual-distribution strategy to capture global and local data characteristics. Despite being originally developed for image tasks, the theoretical foundation and inherent nature of black-box testing enable the HDA approach to transcend its initial scope, making it versatile across various domains, including transformer-based PHM systems involving time-series data. However, its reliance on Variational Autoencoders for latent space approximations fails to account for the temporal dependencies and dynamic variability crucial to time-series data in ICPS, limiting its ability to generate representative adversarial examples. Furthermore, HDA lacks efficient optimization tailored to the time-series domain and does not incorporate physical constraints, reducing its effectiveness in producing realistic and operationally plausible adversarial examples for industrial applications. Therefore, while previous methods are often effective in their specific fields, they are generally time-consuming and frequently overlook the stringent physical constraints associated with industrial time-series data. This limitation leads to low sampling efficiency, as seen with approaches like genetic algorithms, which are inadequate for applications such as PHM systems for PEMFC. To address this, we incorporate the state-of-the-art Artificial Rabbits Optimization (ARO) algorithm to meet the unique requirements of these applications. ARO, which balances exploration and exploitation strategies inspired by natural foraging behavior, efficiently generates physically constrained adversarial examples while significantly outperforming traditional methods in computational efficiency.

In contrast, our approach integrates global and local data insights within a four-tiered hierarchical framework. It begins with Kernel Density Estimation (KDE) combined with a Long Short-Term Memory Variational Autoencoder (LSTMVAE) encoder, which captures temporal relationships in the latent space. This facilitates accurate approximation of global data distributions in time-series data, enabling the effective identification of potential test seeds. Following this, a local robustness indicator is calculated to refine the selection of test seeds by leveraging insights from the global distribution and local characteristics of DL-based PHM system. Subsequent steps involve applying physical constraints to these test seeds based on the features of industrial application, such as PEMFC, ensuring the resulting Adversarial Examples (AEs) are realistic and compliant with physical principles. The generation and optimization of these AEs are then executed through the ARO algorithm, focusing on identifying the most effective adversarial instances. This strategy ensures the discovery of high-quality adversarial samples by harmonizing overall data distributions with specific local information, thereby testing the model's adversarial robustness in a trustworthy way. The contributions of this paper are summarized below:
\begin{enumerate}
    \item This paper presents Hierarchical Testing with Rabbit Optimization (HERO), a novel black-box adversarial testing framework which combines hierarchical data analysis with efficient optimization techniques to identify vulnerabilities in industrial deep learning models through physically constrained examples.
    \item The proposed method integrates the accuracy of hierarchical testing with the efficiency of the state-of-the-art ARO algorithm, optimizing the identification and generation of high-quality time series adversarial samples for robustness evaluation.
    \item Empirical validation of the HERO on real-world dataset, demonstrating its effectiveness in enhancing the resilience of PHM systems and the broader field of ICPS against threats posed by black-box models.
\end{enumerate}


\section{Preliminaries and Related Work}\label{sec:2}
\begin{figure*} [htbp]
  \centering
  \includegraphics[width=\textwidth]{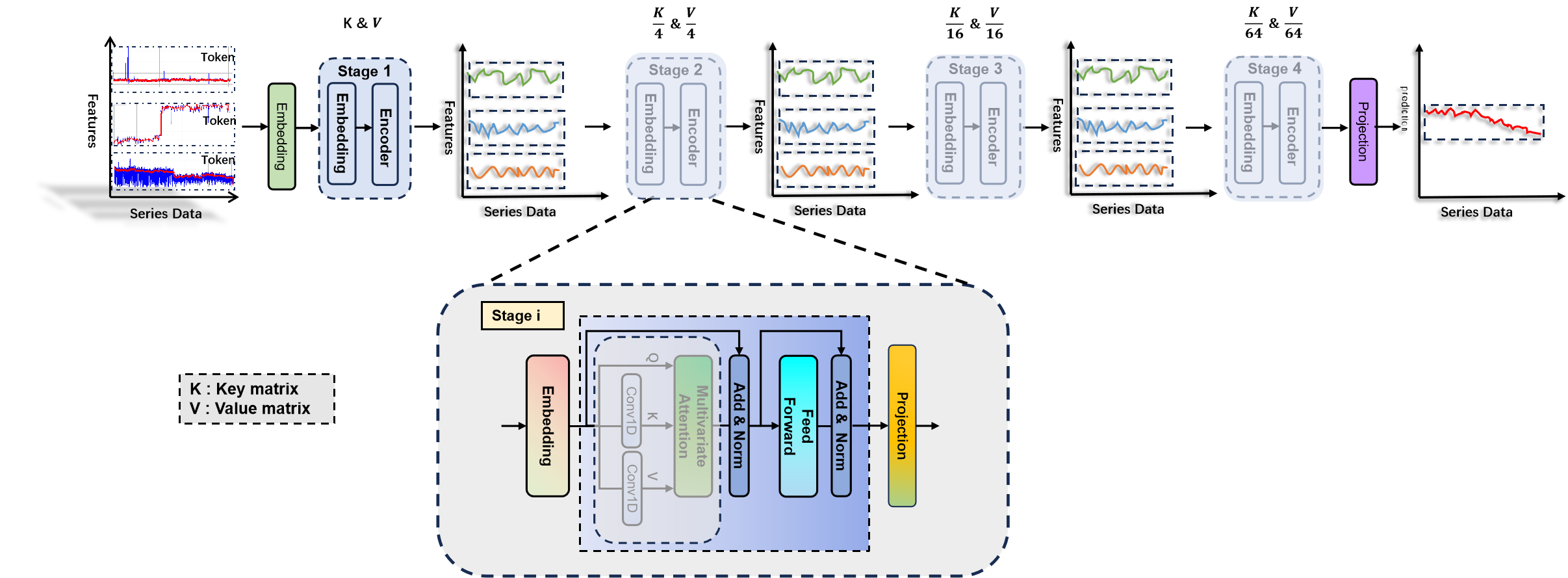}
  \caption{Schematic Overview of the TSTransformer Architecture.}
  \label{fig: Tstransformer structure}
\end{figure*}
\subsection{PHM Systems for Energy RUL Prediction in ICPS}
With the rapid development of ICPS, PEMFCs have become a pivotal energy source, widely integrated into industrial applications such as manufacturing, transportation, and grid stabilization \cite{tellez2021proton}. These applications demand efficient and accurate energy RUL prediction. However, the dynamic operational conditions and complex physical interactions in ICPS present significant challenges. Fluctuating parameters hinder accurate and robust monitoring through PHM systems \cite{10592003, jouin2013prognostics}. Recent methods, such as an LM-BPNN-based power routing approach \cite{10124028} leveraging voltage, current, and temperature data to extend system lifetime, and the VLSTM-LWSAN framework \cite{ZHANG2023108986}, which aligns fine-grained subdomain features in latent space, have shown promise in improving RUL prediction. The 1-DCNN–BiGRU model \cite{9992207} has made progress in capturing spatial and temporal features by integrating 1-D convolutional layers for spatial feature extraction and BiGRU for temporal feature extraction, effectively advancing feature fusion in spatiotemporal dimensions. Despite these advancements, their limitations in handling multivariate temporal data and varying temporal scales remain evident. To overcome these challenges, transformer-based architectures have been increasingly adopted in PHM systems, demonstrating superior performance in RUL prediction for industrial battery energy systems by effectively modeling long-term dependencies and complex sequential patterns \cite{lv2023transformer,zhou2023prediction}. 



\subsection{Temporal Scale Transformer Structure}
The Temporal Scale Transformer (TSTransformer) is designed for predicting events from temporal sequences where event timings are irregular \cite{tang2025temporal}.
The TSTransformer model thoughtfully modifies the traditional Transformer architecture by incorporating one-dimensional convolutions that can scale the key ($K$) and value ($V$) matrices into the attention mechanism, which completely changes the analysis of multivariate time series. Furthermore, the inverted Transformer (iTransformer) model embeds the time points of a single sequence into variable labels, and applies feedforward networks (FFN) to each variable token idea. These modifications are specifically tailored to improve the processing of time series characteristics. The architecture is partitioned into a sequence of stages, each consisting of a Transformer encoder that refines the multivariate time series data for RUL prediction in PEMFC systems. Commencing with an embedding layer that encodes input series data into high-dimensional tokens, the process unfolds through spatially scaled one-dimensional convolutions within the successive stages. These stages implement a pyramidal structure, sequentially halving the spatial dimension of the $K$ and $V$ matrices to capture temporal dynamics across various scales. During this hierarchical processing, the embedding of each token is consistently preserved at its original resolution to serve as the Query (Q) matrix, whereas only the Key (K) and Value (V) matrices undergo progressive down-sampling through one-dimensional convolutional (\texttt{Conv1D}) layers. Retaining a full-resolution Q while gradually scaling down the spatial dimensions of K and V is essential for enabling each attention stage to effectively integrate broader contextual information without sacrificing the detailed, fine-grained semantics inherent in the queries. This strategic choice, as visually depicted by the distinct bar widths representing Q, K, and V in Fig.~\ref{fig: Tstransformer structure}, ensures the model simultaneously captures both local temporal patterns and global sequence dependencies efficiently. The convolution scaling ratios for stages 2 to 4 are set at \( \frac{1}{4} \), \( \frac{1}{16} \), and \( \frac{1}{64} \), respectively, enabling the model to discern fine-grained details as well as overarching temporal trends. The final predicted output of the architecture is obtained through a projection layer, which converts the output of the last layer into future series predictions. The framework of the transformer in this paper is illustrated in Fig. \ref{fig: Tstransformer structure}.

\subsection{Model Robustness and Adversarial Examples}
Model robustness refers to the ability of machine learning models to maintain stable prediction performance and decision quality in the face of minor perturbations or changes in input data \cite{jyoti2022robustness}. This attribute reflects the model's adaptability and resilience to the inevitable imperfections in real-world data, such as noise, distortion, or inputs designed with adversarial intent \cite{10263803}. A robust model can preserve its accuracy and reliability under various conditions without experiencing significant performance degradation due to minor anomalies or unforeseen changes in the data. Understanding model robustness allows researchers to systematically enhance a model's resilience against various disturbances through techniques like adversarial training \cite{bai2021recent}, data augmentation \cite{rebuffi2021data}, and regularization \cite{tack2022consistency}. In fields such as ICPS and PHM, ensuring model robustness is crucial, where accurate predictions are essential for system stability and safety in complex and variable environments \cite{torngren2018complexity}. Generally, $f:\mathcal{X}\rightarrow\mathcal{Y}$ be a mapping from the input space $\mathcal{X}$ to the output space $\mathcal{Y}$, representing the PHM model for monitoring lifespan of PEMFC. Then the robustness for time-series model against to the input disturbance can be defined as:
\begin{definition}[Robustness]
A time-series forecasting system is said to be robust if, for all input $x\in\mathcal{X}$ and all disturbance $|d| \leq r$, the following condition holds:
\begin{equation}
    \| f(x+d) - f(x) \|\leq \epsilon
\end{equation}
\end{definition}
where $r$ is the radius of the disturbance, $\|\cdot\|$ denotes the Euclidean norm. Normally, $\epsilon$ is a small positive constant.

Within the conceptual framework of model robustness, a pivotal element in safeguarding the integrity of machine learning models lies in the understanding and mitigation of adversarial examples. These examples are inputs specifically engineered to induce errors in the model's predictions or classifications, exploiting the inherent sensitivity of models to slight, often undetectable alterations in their input data. Such manipulations are subtly designed to mislead the model into yielding erroneous outcomes, all while seeming harmless to human observers. Integral to the efforts of enhancing deep learning robustness, validation, and verification (V\&V) strategies primarily concentrate on the detection and counteraction of these adversarial inputs. This includes the implementation of methods like adversarial attack simulations \cite{goodfellow2014explaining} and coverage-guided testing \cite{9451178}, which are critical for uncovering and addressing the vulnerabilities exposed by adversarial examples, ensuring models remain resilient against such deceptive tactics.

Following the definition of adversarial examples shown in Equation~\ref{AE}, this paper seeks to rigorously test the resilience of state-of-the-art PHM models like the TSTransformer. The aim is to expose these models to minimal perturbations that nevertheless cause significant prediction deviations, simulating realistic operational challenges. Such testing  validates the robustness of these models against both typical and atypical disturbances that could arise from operational errors, equipment malfunctions, or deliberate adversarial attacks \cite{krichen2023formal}. These perturbations might manifest as natural fluctuations \cite{chen2022robust} or as artificially generated AEs \cite{zhao2017generating}. By ensuring that predictions remain accurate and reliable under such conditions, this method confirms the models' effectiveness in real-world applications, thus enhancing PHM strategies for PEMFC systems within the renewable energy sector\cite{zhang2024tabnet}.

\section{Problem Formulation}\label{sec:3}
In PHM systems for PEMFC, accurate prediction of stack voltage is crucial because it significantly impacts the energy efficiency and reliability of the systems \cite{hua2020health}. These systems encounter dynamic variations in operational parameters such as temperature, gas pressure, and flow rates of air or hydrogen, presenting substantial challenges to the design of predictive models \cite{kandlikar2009thermal}. Additionally, natural noise and intentionally crafted adversarial noise can complicate predictions further, potentially leading to disastrous consequences if not properly recognized and addressed. Accurate RUL estimation is essential; premature predictions can result in unnecessary maintenance, whereas delayed predictions might lead to catastrophic failures \cite{9356218}. This balance underscores the need for robust and reliable predictive capabilities in PHM systems or even other ICPS applications.

The primary objective of this study is to test the robustness of predictive models for PEMFC within PHM systems. We aim to develop a black-box adversarial testing framework designed to identify and exploit vulnerabilities in these models. This framework will generate minimal yet highly effective perturbations that significantly impact model predictions, assessing whether the models can withstand real-world operational variations and adversarial threats.

The perturbations are systematically generated to mimic potential real-world disturbances that the models must be capable of handling. These perturbations are formally defined in Equation \ref{AE}. The goal is to minimize the magnitude of these perturbations (\(\delta\)) while ensuring that the deviation in the model's prediction exceeds an acceptable threshold. This approach tests the model's ability to maintain accurate predictions within a defined tolerance under adversarial conditions, thus highlighting critical areas for improving the resilience and reliability of PHM systems in operational environments.

\begin{definition}[Adversarial Examples for Regression Tasks]
For regression tasks such as predicting the RUL of PEMFC systems, more realistic adversarial examples aim to achieve minimal perturbation while causing significant prediction error. These adversarial examples are defined as:
\begin{equation}\label{AE}
    \min_{\delta} \| \delta \| \text{ subject to } |f(x+\delta) - f(x)| \geq \epsilon \text{ and } x+\delta \in \mathcal{X}
\end{equation}
where \(f\) is the regression model, \(x\) is the original input, \(\mathcal{X}\) is the input space, and \(\epsilon\) is a predefined threshold indicating a minimum acceptable level of prediction deviation considered as significant for the system’s operational integrity.
\end{definition}

In this study, the framework is implemented as a black-box method, without requiring knowledge of the internal mechanisms of state-of-the-art predictive PHM models, notably the TSTransformer. This framework assesses the resilience of this kind of models by exposing them to highly effective and realistic test cases. These test cases are generated using the ARO algorithm considering both global data distribution and local data characteristics. The primary objective is to identify vulnerabilities in such advanced PHM models within the ICPS area, thereby evaluating the overall reliability and safety of industrial applications in their operational environments such as PEMFC applications illustrated in this paper.

\section{Methodology}\label{sec:4}
\begin{figure}[htbp]
\centering
	\includegraphics[width=\columnwidth]{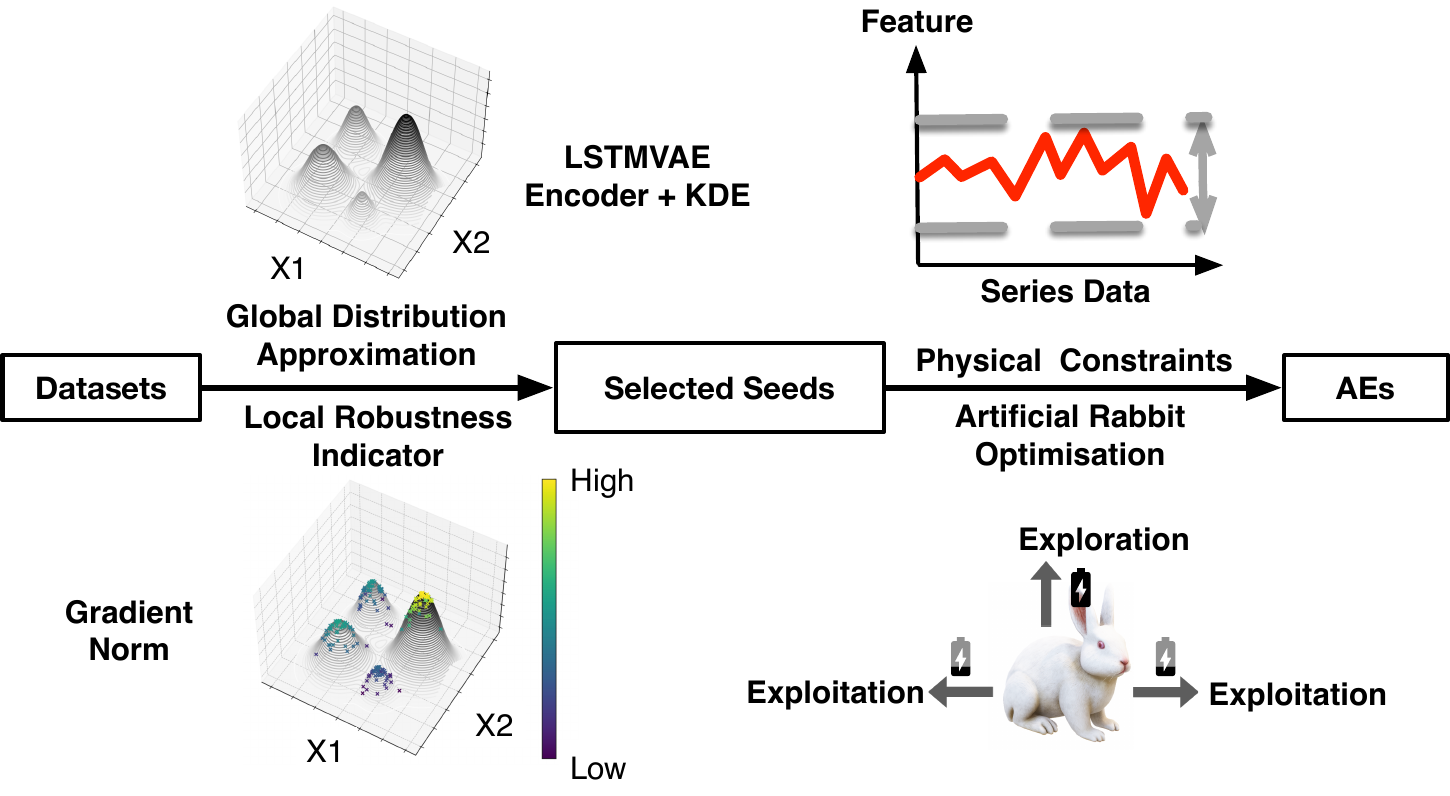}
	\caption{Framework of Hierarchical Testing with Rabbit Optimization}\label{hda}
\end{figure}
In this section, we design a hierarchical testing algorithm shown in Fig.~\ref{hda} to search for adversarial examples with the minimum distance to original data, and thus making them realistic in real-world scenarios. Section \ref{GR-section} explores Global Robustness and Distribution Approximation. Section \ref{LRI} examines Local Robustness. Test seeds selection and cases generation are discussed in \ref{TSS} and \ref{TCG}, respectively. 

\subsection{Global Robustness} \label{GR-section}
To evaluate the global robustness for this task, we integrate local robustness introduced in Section \ref{LRI} with the global probability estimates described in Section \ref{GRA}. The definition of Global Robustness is shown in Definition \ref{GR}.

\begin{definition}[Global Robustness] \label{GR}
The global robustness of a PHM model \( f(x) \) for handling RUL prediction is defined as:
\begin{equation}
R_g := \sum_{n \in X} p_g(x \in n) R_l(n, y, \epsilon)
\end{equation}
where \( p_g(x \in n) \) is the global distribution of region \( n \) (i.e., a pooled probability of all inputs in the region \( n \)), and \( R_l(n, y, \delta) \) is the local robustness of region \( n \) to the target output \( y \) within an acceptable deviation \( \epsilon \).
\end{definition}

\subsubsection{Global Distribution Approximation} \label{GRA}
In current distribution-aware deep learning testing methodologies, reducing the dimension of the input data by Variational Autoencoder (VAE) emerges as a key initial step to efficiently processing and analyzing data\cite{dola2021distribution, berend2021distribution}. In this paper, by applying the LSTMVAE model to a preprocessed dataset \(X = \{x_1, x_2, ..., x_n\}\), we achieve not only effective dimensionality reduction but also reveal the data distribution characteristics of the data. This step lays the foundation for in-depth data analysis, which, when combined with KDE, estimates Probability Density Function (PDF) through representations in the latent space, thereby offering a deeper understanding of the global distribution of the dataset \cite{10121016}. Here, we assume that the training dataset is unbiased sampled from the real world, which means they have same distribution.

The LSTMVAE-encoder $f_{\text{Encoder}}$ processes each input sequence \(x_i\) to capture temporal dependencies, leading to a set of hidden states. The last hidden state \(h_{i,\text{final}}\) is then used to compute the distribution of the latent space:
\begin{equation}
(\mu_i, \log\sigma_i^2) = f_{\text{Encoder}}(h_{i,\text{final}})
\end{equation}
Here, \(\mu\) represents the mean vector and \(\log\sigma^2\) represents the log variance vector of the latent space. These parameters reveals the distribution nature of the latent space where the data's global characteristics are captured.

To enable the gradient-based optimization of the model, the reparameterization trick is employed, allowing the model to sample from the latent distribution in a differentiable manner \cite{course2024amortized}:
\begin{equation} \label{latent}
z_i = \mu_i + \exp\left(\frac{\log\sigma_i^2}{2}\right) \cdot \epsilon_i, \quad \epsilon_i \sim \mathcal{N}(0, I)
\end{equation}
The Equation \ref{latent} defines the sampled latent variable \(z_i\) for each input sequence \(x_i\), where \(\epsilon_i\) is an auxiliary noise vector sampled from a standard normal distribution, ensuring the stochasticity of \(z_i\). \(\mathcal{N}(0, I)\) represents the standard normal distribution with a mean of 0 and a covariance matrix equal to the identity matrix \(I\), serving as the prior distribution for the latent variables.

The loss function for our LSTMVAE, which is minimized during training, is comprized of two terms: the reconstruction error and the Kullback-Leibler (KL) divergence. The KL divergence acts as a regularizer by measuring the divergence of the latent distribution from a standard normal distribution. The overall loss function is given by:
\begin{equation}\nonumber
\begin{split}
\mathcal{L}(x_i, \mu_i, \log\sigma_i^2) &= \text{MSE}(x_i, \hat{x}_i) \\
&\quad + \beta \cdot D_{KL}(\mathcal{N}(\mu_i, \exp(\log\sigma_i^2)) \,||\, \mathcal{N}(0, I))
\end{split}
\end{equation}
where \(\text{MSE}\) represents the mean squared error, a measure of the reconstruction error between the original data \(x_i\) and the reconstructed data \(\hat{x}_i\). \(D_{KL}(\cdot)\) is the KL divergence, quantifying the deviation of the learned latent distribution, parameterized by \(\mu_i\) and \(\sigma_i^2\), from the standard normal distribution \(\mathcal{N}(0, I)\). This term acts as a regularizer, ensuring that the latent space distribution does not deviate significantly from a standard normal distribution. \(\beta\) is a hyperparameter that balances the reconstruction error and the KL divergence. 

Following the calculation of $\mu$, $\sigma^2$ of the training data, KDE is utilized as a non-parametric method to estimate the PDF of these variables \cite{kang2018development}. Regarding the outputs of LSTMVAE-encoder, \(\mu_i\) represents the mean and \(\log\sigma_i^2\) calculates the variance of the distribution for each input sequence in the latent space. To perform KDE, the means \(\mu_i\) from the LSTMVAE-encoder are considered as data points in this latent space which are used in Equation \ref{KDE}. KDE helps to create a continuous PDF that shows the global distribution of the data points, effectively mapping out where data points are most densely clustered in the latent space. This approach provides a view of the data's distribution in a simplified and continuous form. The KDE in the latent space can be formulated as:
\begin{equation} \label{KDE}
\hat{k}(z) = \frac{1}{N} \sum_{i=1}^{N} K_{h_i}\left(z - \mu_i\right)
\end{equation}

where \(\hat{k}_(z)\) is the estimated probability density in the latent space at point \(z\). \(z\) represents points in the latent space. \(N\) is the number of observations in the time series. \(K\) is the kernel, a non-negative function that integrates to one and has a mean of zero. Our choices for \(K\) include the Gaussian function, Epanechnikov function and Exponential function. \(h_i\) = $\sigma_i$ is the bandwidth parameter that controls the width of the kernel and influences the smoothness of the resulting density estimate. We empirically use \(\hat{k}(z)\) to approximate the global distribution which is introduced in \cite{huang2023hierarchical}, denoted as \(g(z)\). This approximation is then utilized in Definition \ref{GR} to compute the global robustness.

\subsection{Local Robustness Indicator} \label{LRI}

In regression tasks, local robustness measures the model’s ability to maintain consistent predictions within a specified tolerance around a target output for all inputs in a small neighborhood. Unlike classification tasks, where predictions must match a specific label, regression tasks require that the model’s predictions do not deviate beyond an acceptable error margin from the true continuous values. The formal DL robustness definition is reused from \cite{webb2018statistical} to ensure generality.
\begin{definition}[Local Robustness]
For a PHM model \( f(x) \) handling RUL prediction, local robustness in a region \( \eta \) with respect to a target output \( y \) and an acceptable deviation \( \epsilon \) is defined as:
\begin{equation}
R_l(\eta, y, \epsilon) := \int_{x \in \eta} I(x) p_l(x | x \in \eta) dx 
\end{equation}
\end{definition}

where \( I(x) = 1 \) if \( |f(x) - y| \leq \epsilon \) (the prediction is within the error margin), and \( I(x) = 0 \) otherwise. The function \( p_l(x | x \in \eta) \) represents the probability distribution of \( x \) within the region \( \eta \), focusing on how inputs are dispersed or expected within this neighborhood.

To better understand the local robustness of models trained on time series data of PEMFC with TSTransformer, we propose shifting away from traditional sampling techniques due to their high resource demands and limited accuracy in reflecting model behavior near specific inputs. Instead, we propose using gradient computations with respect to the input data as practical Auxiliary Information, which has strong correlations with local robustness $R_l$. This information is termed the Local Robustness Indicator (LRI). Given a PHM model \(f\), input data sequence \(x\), and corresponding outputs \(y\), the computed gradient norm (GN) is defined as: $\mathcal{G} = \| \nabla_{\text{x}} \mathcal{L}(f(x), y) \|_{\infty}$.
This measure reflects the highest rate of increase of the loss function with respect to changes in the input data, thereby indicating the model's local vulnerability and sensitivity. A higher LRI suggests that small perturbations in the input data could lead to significant increases in the loss, highlighting areas where the model may be less robust.

\subsection{Test Seeds Selection} \label{TSS}

Test seeds are selected by normalizing both the local robustness indicators and the global distribution values to a common scale. This normalization process allows us to synthesize a ranking metric that incorporates both local and global characteristics of the predictive model by leveraging the insights gained from the global distribution approximations discussed in Section \ref{GRA} and the gradient analysis introduced in Section \ref{LRI}.
This method systematically identifies the test seeds that are most likely to reveal potential vulnerabilities within the model, thereby enabling a more efficient and focused approach to selecting optimal seeds for testing. The formula for the test seeds selection ranking is based on approximated global robustness $\hat{R_g} = Nor(\mathcal{G}) \cdot Nor(g(z))$.
With this scores, we then select the top-$k$ test cases as our test seeds, where $k$ is determined by the available testing budget.

\subsection{Physical Constraints on Features} \label{constraints}
Before generating test cases, incorporating physical constraints into the generation process of AEs is crucial to ensure these examples are realistic and adhere to real-world operational scenarios \cite{song2018physical}. This is particularly important in PHM systems for PEMFC, where operational parameters must be maintained within specific limits to ensure system integrity.

To effectively apply these physical constraints, we introduce a constraint function \(C\) that enforces generated test cases which are detailed in Section \ref{AROTCG} to remain within their physically allowable limits. This function ensures that each feature of the newly generated cases does not exceed its specified operational threshold, thus maintaining the realism of the generated AEs for PEMFC. The constraint function \(C\) is applied directly to the features of the new data points during optimization process, ensuring they adhere to the upper and lower bounds defined by the physical constraints of the real-world system. The constraint function can be mathematically represented as:
\begin{equation}
C(\tilde{x}_t^i) = \begin{cases} 
  l_i & \text{if } \tilde{x}_t^i < l_i, \\
  u_i & \text{if } \tilde{x}_t^i > u_i, \\
  \tilde{x}_t^i & \text{otherwise},
\end{cases}
\end{equation}

where \(\tilde{x}_t^i\) represents the feature of newly generated data from the test seed at timestep $t$, and \(l_i\) and \(u_i\) denote the lower and upper bounds for feature \(i\), respectively. This ensures that the feature \(x'_i\) is adjusted to \(l_i\) if it falls below the lower bound, adjusted to \(u_i\) if it exceeds the upper bound, or remains unchanged if it is within the bounds.

By applying this constraint function to the newly generated data during the test case generation phase, we ensure that the AEs produced are within the feasible operational range of the system's parameters. This method preserves the integrity of the PHM system's performance assessment and ensures that the testing reflects conditions that the system is likely to encounter in actual use, enhancing the evaluation's reliability.

\subsection{Test Cases Generation} \label{TCG}
\subsubsection{Optimization Problem Formulation} \label{OPF}
For testing the robustness of PHM systems in ICPS, our objective is to generate adversarial examples that maximize the predictive error while remaining statistically plausible within the input data distribution. The optimization problem for a tuple $(x_{\textit{seed}}, y_{\textit{seed}})$ can be formulated as follows:
\begin{equation}
\begin{aligned}
\text{maximize} \quad & \mathcal{L}(f(\tilde{x}_t), y_t) + \alpha \cdot \mathcal{L}(\tilde{x}_t, x_t) \\
\text{subject to} \quad & |\tilde{x}_t^i - x_t^i| \leq r^{i}_{c}, \quad \text{for all } i = 1, 2, \ldots, j
\end{aligned}
\end{equation}
where $\mathcal{L}(f(\tilde{x}_t), y_t)$ is the loss function assessing prediction error for perturbed input $\tilde{x}_t$ compared to the actual output $y_t$, aiming to maximize model prediction discrepancies. $\mathcal{L}(\tilde{x}_t, x_t)$ measures the perturbed input's deviation from the original, ensuring perturbations are realistic within the data distribution. The coefficient \(\alpha\) balances prediction error maximization with input fidelity, while \(r^{i}_{c}\) defines the maximum permissible change for each feature \(i\) (out of \(j\) features) which is calculated by constraint function $C$, ensuring all perturbations are subtle yet effective in industrial context.

\subsubsection{Convergence Analysis}
The convergence properties of the ARO algorithm, as a meta-heuristic algorithm, satisfy Hypothesis 1 of the Global Search Convergence Theorem \cite{solis1981minimization}. Hypothesis 1 states that \textit{the objective function sequence must be non-increasing over generations, ensuring that each subsequent generation either improves or maintains the quality of the solution}. Based on this theoretical foundation, the ARO algorithm is designed to optimize time-series adversarial examples \(\tilde{x}_t\) in accordance with Hypothesis 2 \cite{solis1981minimization, zhao2020artificial}, which can be summarized as \textit{the zero probability of consistently missing any positive-volume subset \(B\) of the search space \(S\)}.
\begin{equation}
\prod_{g=0}^\infty \left[ 1 - \mu_g(B) \right] = 0
\end{equation}
This implies that for any adversarial subset \(B \subseteq S\) with positive volume in our optimization presented in \ref{OPF}, the probability of consistently missing \(B\) is zero. The probability measure \(\mu_g(B)\) represents the likelihood that the \(g\)-th generation of ARO covers the subset \(B\), thus satisfying the global convergence criterion. Therefore, as the number of generations approaches infinity, the probability measure of ARO can be expressed as:
\begin{equation}
\lim_{g \to \infty} P(\tilde{x}_g \in Opt_{\xi, M}) = 1
\end{equation}
where \(\{\tilde{x}_g\}_{a=0}^{\infty}\) represents the sequence of adversarial examples produced by ARO over infinite generations. The region \( Opt_{\xi, M} \) is the optimality region, encompassing solutions with an objective function value within \(\xi\) of the optimum, given \(M\) as the allowable margin of deviation. This indicates that, as \(g\) becomes very large, the probability of the solution falling within this optimality region approaches one, confirming that ARO converges to the global optimal adversarial example, effectively testing the robustness of the PHM system. 

\subsubsection{Implementation Details} \label{AROTCG}
To generate AEs that adhere to the physical constraints established in the Section \ref{constraints}, we employ the concepts of ARO algorithm \cite{wang2022artificial}. The ARO algorithm is inspired by the natural foraging behavior of rabbits, which embodies the principles of exploration (Detour foraging) and exploitation (Random hiding) to locate the optimal food sources. In the context of generating AEs, the ARO algorithm simulates a population of 'rabbits' (test seeds) exploring the feature space to find the most challenging inputs for the model under test. The algorithm dynamically adjusts its search strategy based on an energy factor, which represents the algorithm's tendency to explore new areas of the search space or exploit known regions to refine existing solutions. This balance is crucial for avoiding local optima and ensuring a comprehensive search. 

The generation of adversarial examples is guided by a composite loss function \( L \), designed to ensure that while the adversarial examples are effective in exposing vulnerabilities, they remain realistic and representative of genuine operational scenarios. The function \( L \) is composed of two main components: \( L_{\text{pred}} \) and \( L_{\text{sim}} \).

Predictive loss \( L_{\text{pred}} \) is calculated as the sum of MSE across the forecast horizon, quantifying the accuracy of the model's predictions against the actual observed RUL (\( \mathbf{Y_{\text{seed}}} \)) \cite{liu2019remaining, hua2021remaining, song2024data}. This metric is crucial for capturing the temporal dynamics associated with RUL predictions:
\begin{equation}
L_{\text{pred}} = \sum_{t=1}^{T} \left( y_{\textit{pred}, t} - y_{\textit{seed}, t} \right)^2
\end{equation}
Here, \( T \) represents the total number of time steps, and \( y_{\textit{pred}, t} \) and \( y_{\textit{seed}, t} \) are the predicted and actual values at each time step, respectively.

Similarity Loss \( L_{\text{sim}} \) normally evaluates how closely the generated examples resemble the original inputs (\( \mathbf{X_{\text{seed}}} \)) \cite{pei2021towards, 9140397, brophy2023generative}. It is computed as the average MSE across all dimensions and time steps of the input series, assessing the fidelity of the adversarial examples in replicating the original data behavior:
\begin{equation}
L_{\text{sim}} = \frac{1}{T} \sum_{t=1}^{T} \left( x_{\text{adv}, t} - x_{\text{seed}, t} \right)^2
\end{equation}
where \( x_{\text{adv}, t} \) and \( x_{\text{seed}, t} \) are the adversarial and original inputs at each time step, respectively.

This composite loss function ensures that the AEs are not only challenging for the model but also maintain the integrity and realism necessary for practical applications in real-world ICPS scenarios. The parameter \( \alpha \) serves as a tuning factor to strike a balance between maximizing prediction deviations and preserving the authenticity of the generated samples. This balance is crucial for ensuring that adversarial perturbations are effective while remaining realistic. Additionally, both metrics undergo min-max normalization $Norm( \cdot )$ to align them on a consistent scale, enabling a balanced analysis:
\begin{equation}
    L = \alpha \times Norm(L_{\text{pred}}) - (1 - \alpha) \times Norm(L_{\text{sim}})
\end{equation}
Ultimately, test cases that yield high Prediction Loss while maintaining low Similarity Loss are selected to test the robustness of the model. This approach ensures that the adversarial examples are not only effective in challenging the model but also maintain the integrity and realism of the test scenarios, crucial for their applicability in real-world conditions. The specific steps of using ARO to generate AEs are shown in Algorithm \ref{ARO}:
\begin{algorithm}[htbp]
\caption{Artificial Rabbits Optimisation Algorithm for Generating Adversarial Examples}
\begin{algorithmic}[1]
\State \textbf{Input:} $x_{\text{seed}}$ (test seed), $y_{\text{seed}}$ (expected output), $f$ (model), $G$ (number of generations), $N_{\text{pop}}$ (population size), $\epsilon$ (perturbation limit), $\alpha$ (weighting factor for loss components), $\beta$ (Balance factor for exploration and exploitation)
\State \textbf{Output:} $\tilde{x}_{\text{best}}$ (generated best adversarial example)

\Procedure{ARO}{$x_{\text{seed}}, y_{\text{seed}}, f, G, N_{\text{pop}}, \epsilon, \alpha$}
    \State Initialize population \(\mathbf{Pop}\) of size \(N_{\text{pop}}\) by perturbing \(x_{\text{seed}}\) within bounds \(\epsilon\)
    \For{$g = 1$ to $G$}
        \State Evaluate fitness using $L$ and $\alpha$
        \State $\tilde{x}_{\text{best}} \gets$ individual in $\mathbf{Pop}$ with highest fitness
        \For{each $\tilde{x}$ in $\mathbf{Pop}$}
            \State Calculate energy factor $A$ based on $g$
            \If{$A > \beta$}
                \State $\tilde{x} \gets$ Detour Foraging (Exploration)
            \Else{}
                \State $\tilde{x} \gets$ Random Hiding (Exploitation)
            \EndIf
            \State Apply perturbation
            \State Clamp $\tilde{x}$ within $[x_{\text{seed}} - \epsilon, x_{\text{seed}} + \epsilon]$
            \State Clamp $\tilde{x}_t^i$ using $C(\tilde{x}_t^i)$
        \EndFor
    \EndFor
    \State $\tilde{x}_{\text{best}} \gets$ individual in $\mathbf{Pop}$ with highest final fitness
\EndProcedure
\end{algorithmic}
\label{ARO}
\end{algorithm}

The ARO algorithm is specifically tailored to meet the unique requirements of generating adversarial examples for PHM systems in PEMFC. Recognizing the critical nature of physical constraints in this domain, the algorithm incorporates a mechanism to ensure all generated AEs remain within the operationally plausible bounds, as established in Section \ref{constraints}. This adaptation involves a clamping step within the algorithm's iteration process, which rigorously applies the defined physical constraints to each candidate solution. Moreover, the introduction of diversity at specified intervals helps mitigate the risk of convergence to suboptimal adversarial examples, thereby enhancing the robustness of the generated AEs against the model. These modifications make the ARO algorithm a powerful tool for identifying potential vulnerabilities in PHM systems, ensuring that the model's predictions are reliable under a wide range of realistic conditions.



\section{Results and evaluation}\label{sec:5}
\subsection{Experiment Settings}
\begin{figure} [htbp]
  \centering
  \includegraphics[width=0.8\columnwidth]{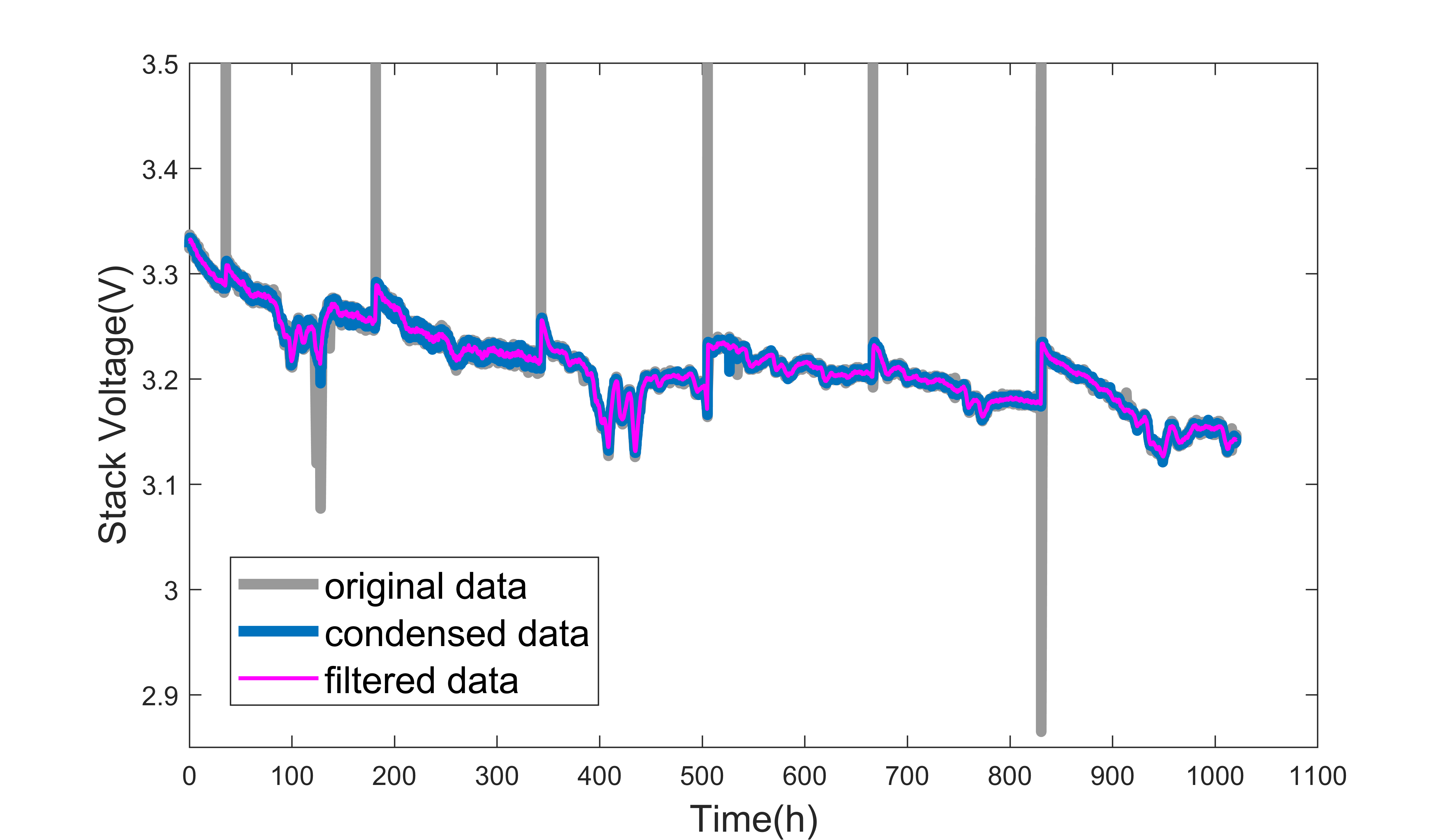}
  \caption{Visualisation of original data, condensed data and filtered data.}
  \label{fig:data preprocessing}
\end{figure}

\paragraph{Datasets and Preprocessing} 
The test data in this study is sourced from the IEEE PHM 2014 Data Challenge quasi-dynamic dataset \cite{gouriveau2014ieee,liu2018short,kim2016online}, collected using a specialized fuel cell (FC) test bench with a power capacity of up to 1 kW. The details of each feature is shown in Table~\ref{tab:features}. The durability test spans 127,370 data points over a period of 0 to 1020 hours. To manage the dataset size, data points were recorded every six minutes, reducing the dataset to 10,134 points. A moving average filter \cite{smith2013digital} was then applied to the condensed dataset to reduce noise while preserving the key characteristics of the raw data, resulting in the final filtered dataset. Operational variables, such as stack voltage, inlet/outlet temperatures, and current, were analyzed, and the most relevant features were selected for RUL prediction. Outliers and distorted data were removed to ensure data quality. The statistical properties of the raw, condensed, and filtered datasets are visualized in Fig.~\ref{fig:data preprocessing} and summarized in Table~\ref{tab:stat_comparison}, showing minimal differences in metrics such as mean, standard deviation, and range. This confirms that the preprocessing steps preserved the physical characteristics of the data while ensuring alignment with real-world distributions.

Before being input into the PHM model, all variables were normalized to a smaller range to facilitate training. To maintain consistency with real-world data distributions, adversarial examples generated using HERO were clamped to the normalized range using a predefined constraint function $C$, ensuring that all variables remained within operational limits and data integrity was preserved during adversarial example generation.
\begin{table}[ht]
\centering
\caption{Health Monitoring Features for PHM Models}
\label{tab:features}
\begin{tabular}{ll}
\hline
\textbf{Features} & \textbf{Explanations} \\
\hline
Time (h) & aging time  \\
U1-U5, Utot (V) &  Five single cell voltage and stack voltage \\
I (A), J (A/cm\textsuperscript{2}) & Current and current density  \\
TinH2, ToutH2 (°C) & The inlet and outlet hydrogen gas temp. \\
TinAIR, ToutAIR (°C) & The Inlet and outlet air temp. \\
TinWAT, ToutWAT (°C) & The Inlet and outlet cooling water temp. \\
PinH2, PoutH2 (mbara) & The inlet and outlet hydrogen gas pressure \\
PinAIR, PoutAIR (mbara) & The inlet and outlet air pressure \\
DinH2, DoutH2 (L/min) & The inlet and outlet hydrogen gas flow rate \\
DinAIR, DoutAIR (L/min) & The inlet and outlet air flow rate \\
DWAT (l/mn) & The cooling water flow rate \\
HrAIRFC (\%) & The estimated hygrometry of inlet air \\
\hline
\end{tabular}
\end{table}

\begin{table}[htbp]
    \centering
    \caption{Statistical Summary for Physical Constraints}
    \label{tab:stat_comparison}
    \begin{tabular}{lccc|c}
        \toprule
        \textbf{Variable} & \textbf{Dataset} & \textbf{Mean} & \textbf{Std Dev} & \textbf{Constraint Range} \\
        \midrule
         \( U_{\text{tot}} \) & Raw       & 27.276 & 1.968   & [-2.7427, 2.5039] \\
                                             & Filtered    & 27.378 & 1.938  & \\
        \midrule
         \( T_{\text{inH2}} \) & Raw       & 45.00  & 1.429   & [-0.1266, 1.8589] \\
                                                     & Filtered    & 45.07  & 1.415   & \\
        \midrule
         \( T_{\text{inAIR}} \) & Raw      & 38.672 & 0.642   & [0.5131, 7.7111] \\
                                                & Filtered    & 38.705 & 0.636  & \\
        \midrule
         \( T_{\text{outH2}} \) & Raw      & 53.701 & 0.076   & [-4.0251, 3.5622] \\
                                                       & Filtered    & 53.699 & 0.076   & \\
        \midrule
         \( T_{\text{inWAT}} \) & Raw       & 70.069 & 0.052  & [-13.6818, 9.8199] \\
                                                  & Filtered    & 70.070 & 0.050   & \\
        \midrule
        \( I \) & Raw      & 1267.2 & 1.901   & [-3.1135, 3.4389] \\
                          & Filtered    & 1267.3 & 1.891   & \\
        \midrule
         \( P_{\text{outAIR}} \) & Raw       & 52.308 & 0.210   & [-2.3068, 2.6617] \\
                                                      & Filtered    & 52.315 & 0.210   & \\
        \midrule
        \( Hr_{\text{AIRFC}} \) & Raw      & 51.265 & 0.107   & [-3.0550, 1.4828] \\
                                                     & Filtered    & 51.265 & 0.109  & \\
        \midrule
         \( T_{\text{outAIR}} \) & Raw      & 3.217  & 0.042   & [-2.8681, -0.0476] \\
                                                  & Filtered    & 3.214  & 0.040   & \\
        \bottomrule
    \end{tabular}
\end{table}

\subsection{Case 1 (PHM Model Effectiveness)}
To validate the effectiveness of PHM systems under normal operating conditions, we first evaluate the performance of various deep learning-based architectures. This case implements a comprehensive evaluation framework with two essential metrics: the percentage error of forecast (\%Er\textsubscript{FT}) and the accuracy score (A\textsubscript{FT})\cite{mao_jackson_2016}. For the RUL estimation, multiple voltage loss thresholds (3.5\%, 4.0\%, 4.5\%, 5.0\%, and 5.5\%) are established to characterize different degradation states of the fuel cell system\cite{sun2023improved}. These thresholds are calculated based on the initial voltage of 3.325V, which serves as the reference point for monitoring voltage degradation. The \%Er\textsubscript{FT} metric quantifies the deviation between predicted and actual RUL values, where positive values indicate early forecasts and negative values represent delayed predictions. The metric is calculated as:

\begin{equation}
\%Er_{FT} = \frac{RUL_{true} - RUL_{prognostic}}{RUL_{true}} \times 100\%
\end{equation}

The accuracy score A\textsubscript{FT} is defined as:
\begin{equation}
A_{FT} = 
  \begin{cases} 
   \exp(-\ln(0.5) \cdot (\%Er_{FT}/5)) & \text{if } \%Er_{FT} \leq 0 \\
   \exp(\ln(0.5) \cdot (\%Er_{FT}/20)) & \text{if } \%Er_{FT} > 0
  \end{cases}
\end{equation}

The final RUL assessment score is computed as:
\begin{equation}
Score_{RUL} = \frac{1}{5} \sum_{} A_{FT},FT \in \{3.5\%,4\%, 4.5\%, 5\%, 5.5\%\}
\end{equation}

As shown in Table~\ref{tab:Comparison result with different models}, comparative experiments demonstrate the superior performance of TSTransformer under normal operating conditions. The model achieves a remarkable Score\textsubscript{RUL} of 0.914, which significantly outperforms traditional deep learning architectures including LSTM (0.463) \cite{liu2019remaining}, Transformer (0.013) \cite{lv2023transformer}, and iTransformer (0.888) \cite{liu2024itransformer}. Additionally, TSTransformer exhibits exceptional precision with the lowest RMSE of 0.003, compared to LSTM (0.022), Transformer (0.025), and iTransformer (0.008). These quantitative results validate the effectiveness of TSTransformer in capturing both short-term dynamics and long-term degradation patterns of PEMFC systems under non-adversarial environments.
\begin{table}[ht]
\setcounter{table}{2}
\renewcommand{\arraystretch}{0.9} 
\setlength{\tabcolsep}{4pt} 
\centering
\caption{Performance Comparison of PHM Systems with Different DL Architectures}
\label{tab:Comparison result with different models}
\begin{tabular}{lcccc}
\toprule
\textbf{Metric} & \textbf{LSTM} & \textbf{Transformer} & \textbf{iTransformer} & \textbf{TSTransformer} \\
\midrule
$Score_{RUL}$ & 0.463 & 0.013 & 0.888 & \textbf{0.914} \\
$RMSE$ & 0.022 & 0.025 & 0.008 & \textbf{0.003} \\
\bottomrule
\end{tabular}
\end{table}

\begin{figure} [htbp]
  \centering
  \includegraphics[width=0.5\textwidth]{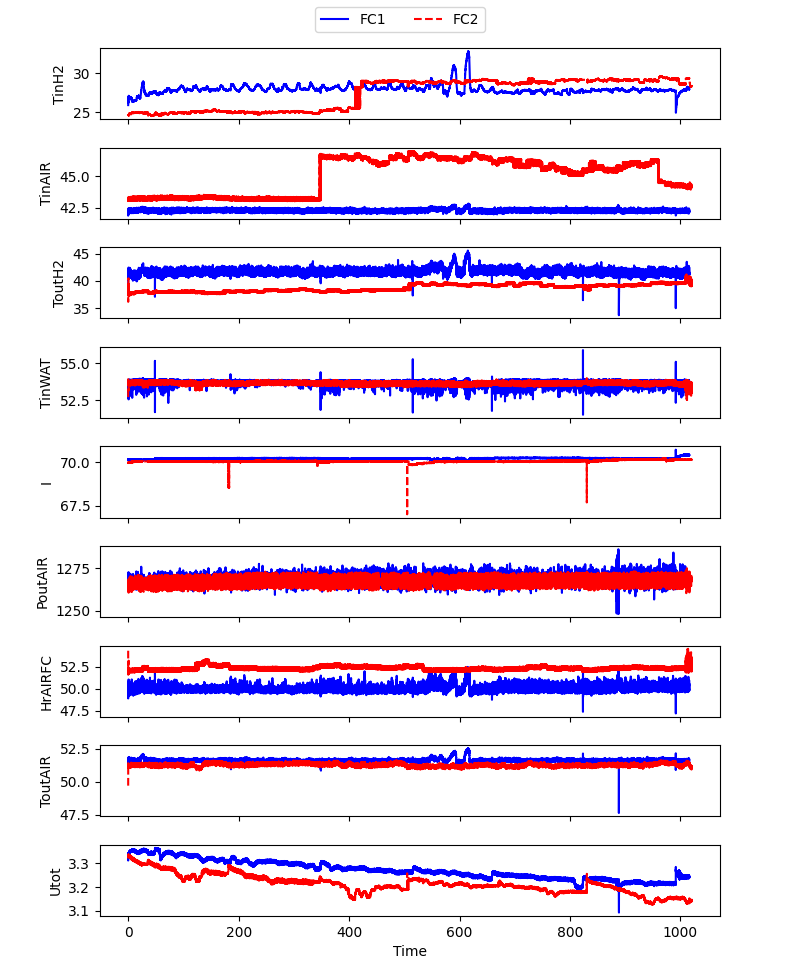}
  \caption{Feature comparsion between FC1 and FC2}
  \label{fig:case1}
\end{figure}

\begin{figure*}[ht]
\centering
\begin{subfigure}{0.5\textwidth}
    \includegraphics[width=\linewidth]{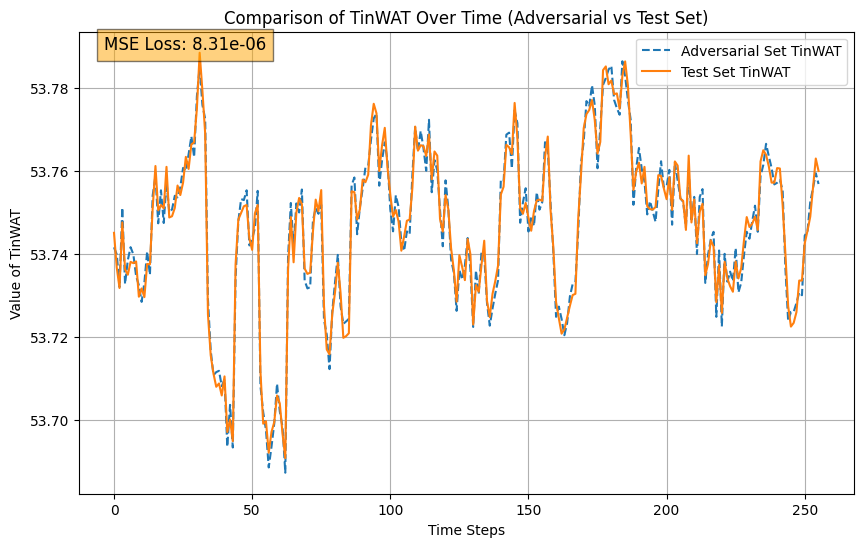} 
    \caption{Comparison of feature-specific values over time}
    \label{fig:adv_feature}
\end{subfigure}%
\begin{subfigure}{0.5\textwidth}
    \includegraphics[width=\linewidth]{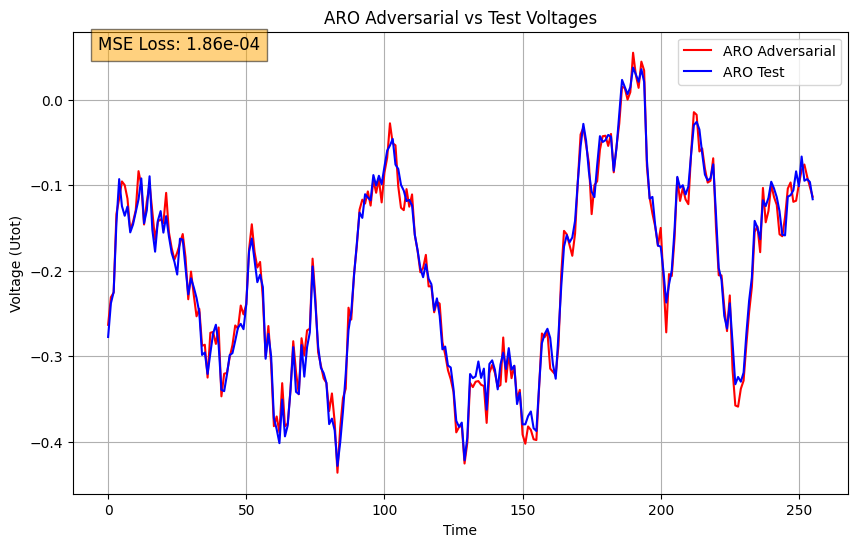} 
    \caption{Comparison of average feature values over time}
    \label{fig:mean_loss}
\end{subfigure}
\caption{Similarity comparison between the adversarial example and test seed.}
\label{fig:combined_figures}
\end{figure*}
\subsection{Case 2 (Data Variability Analysis)}
This section highlights the necessity of testing PHM systems under varied conditions by comparing static and dynamic datasets. Two aging tests were conducted: a stationary regime test (FC1) under nominal operating conditions and a dynamic current test (FC2) with high-frequency current ripples. As summarized in Table~\ref{tab:features}, the datasets include monitoring parameters such as power loads, temperatures, and hydrogen and air stoichiometry rates. A comparison of the parameter variations over time, shown in Fig.~\ref{fig:case1}, reveals that FC1 data exhibit relatively smooth curves with minimal fluctuations, reflecting stable conditions. In contrast, FC2 data demonstrate significant oscillations, particularly in parameters such as TinH2, ToutH2, and current (I), indicating frequent system perturbations under dynamic conditions. Even parameters like ToutAIR and PoutAIR, which are less affected, show subtle fluctuations in FC2. These results highlight the impact of dynamic test conditions on the stability of PEMFC system parameters.

Such data variability, common in real-world scenarios, introduces challenges to PHM systems. Under dynamic conditions, even slight perturbations—whether natural or adversarial—can distort the degradation trends captured by the model, leading to inaccurate RUL predictions. This underscores the need for PHM systems to ensure robustness against data fluctuations and adversarial noise via adversarial testing. Models like TSTransformer, despite their superior performance under nominal conditions, must be equipped to handle these challenges. 

\subsection{Case 3 (advantages: Testing perspective)}

Prediction errors in PHM systems, even when seemingly minor, can lead to serious operational consequences in PEMFC applications. For instance, Liu et al. \cite{liu2019remaining} reported that a nearly accurate RUL estimate (260h vs. actual 261h) delayed maintenance and eventually can contributed to irreversible damage such as membrane drying and voltage collapse \cite{9205397}. Similarly, Zhang et al. \cite{zhang2024tabnet} demonstrated a case where the PHM model predicted an RUL of 0 at a 4\% fault threshold, while the actual value was 220 hours—leading to a 100\% error and potential risks of premature shutdown or undetected degradation, including hydrogen leakage or stack overheating. In another case, a 95.8\% underestimation near a 4.5\% fault threshold risked missing early signs of hydrogen starvation, increasing the chance of thermal runaway. Overestimation can be equally harmful, as seen in BPNN predictions of 336h RUL versus the actual 261h, resulting in delayed maintenance and degraded performance in systems that demand uninterrupted power supply \cite{liu2019remaining}. These real-world failure cases underscore the catastrophic consequences that even small prediction errors can induce, reinforcing the necessity for robust and resilient predictive models in PHM frameworks.

\begin{figure}[ht]
\centering
\includegraphics[width=\linewidth]{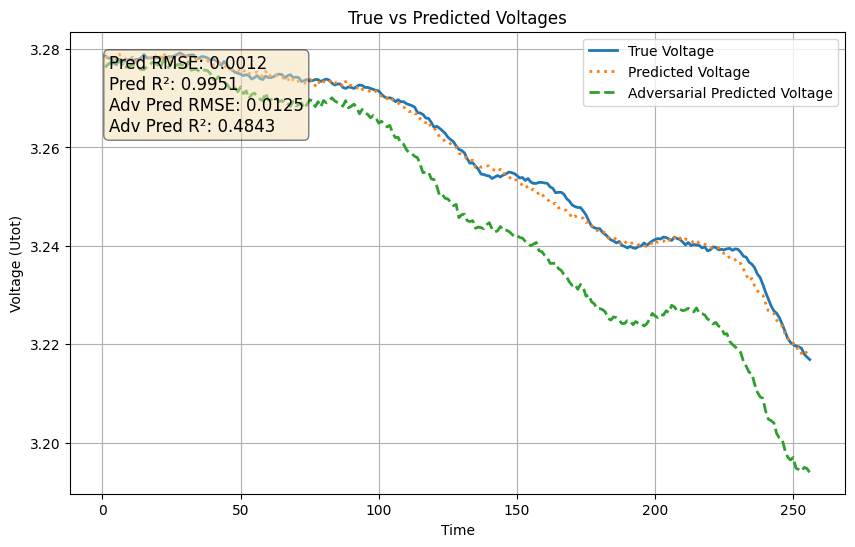}
\caption{Demonstration of significant performance reduction due to adversarial perturbations.}
\label{fig:prediction_loss}
\end{figure}

Therefore, building on the analysis of catastrophic consequences of PEMFC failures, we aim to assess and demonstrate the robustness of deep predictive models within PHM systems against adversarial attacks and dynamic scenarios, our experimental results, as showcased in Fig.~\ref{fig:adv_feature} and~\ref{fig:mean_loss}, meticulously illustrate how minimal perturbations ($\epsilon = 0.03$) can significantly reduce predictive accuracy. For instance, in Fig.~\ref{fig:prediction_loss}, adversarial predictions caused a sharp decline in \( R^2 \) from 0.9951 to 0.4843 and a tenfold RMSE increase from 0.0012 to 0.0125, highlighting the adverse impact on predictive performance. This dramatic contrast highlights the current models' vulnerability to disturbances that are strategically crafted to mimic real-world dynamic conditions, thus underscoring a critical weakness in their robustness under dynamic and adversarial scenarios. These findings emphasize the need for enhanced robustness within PHM systems, particularly in the context of RUL prediction, to safeguard against unforeseen failures and ensure system reliability. The HERO methodology, through its comprehensive testing and validation, demonstrates significant progress in understanding and improving the resilience of such predictive models.

\begin{figure}[ht]
  \centering
  \includegraphics[width=0.5\textwidth]{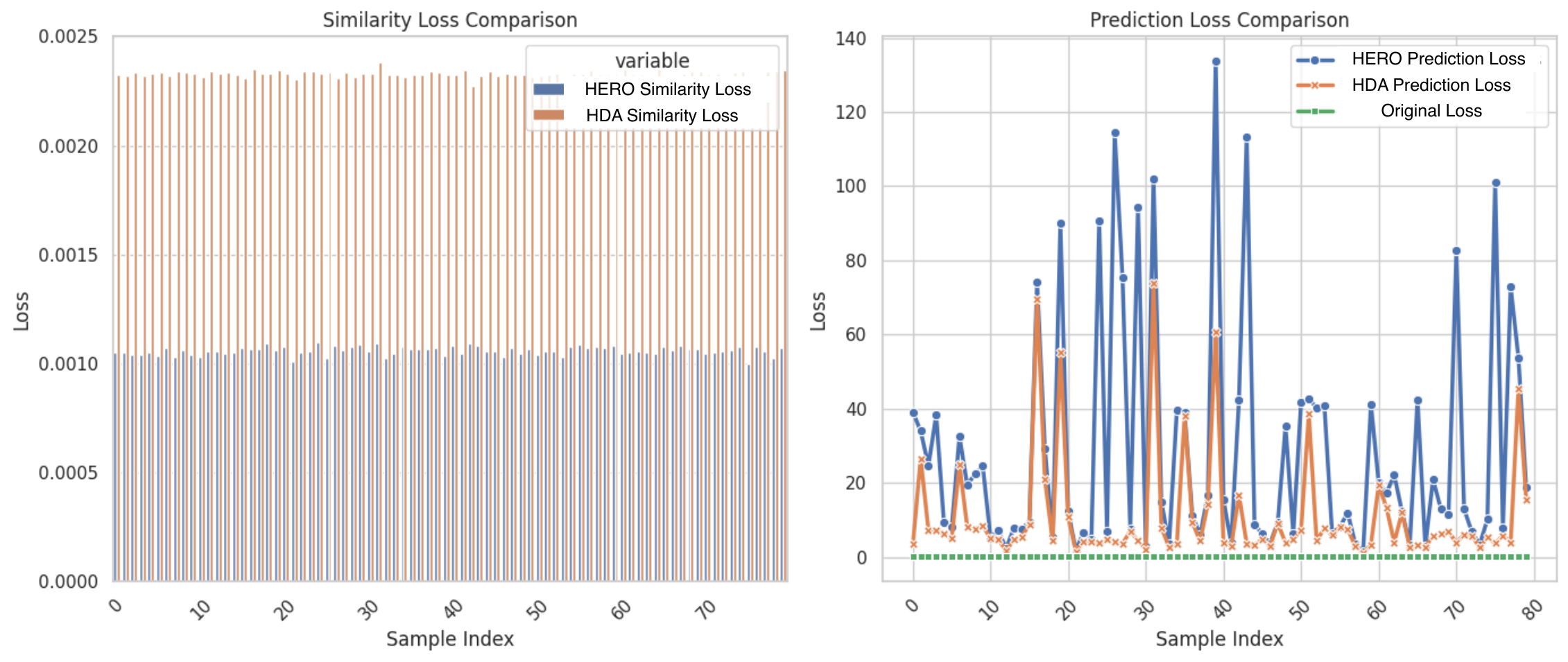}
  \caption{Comparative Analysis between HERO and HDA \cite{huang2023hierarchical}}
  \label{fig:compare_hda}
\end{figure}

Following this foundational analysis, we present a comparative evaluation between our HERO methodology and the state-of-the-art HDA approach, as detailed by Huang et al. \cite{huang2023hierarchical}. Fig.~\ref{fig:compare_hda} provides a detailed comparison using data from 80 randomly selected samples across numerous trials with identical predefined conditions, where each data point represents an individual sample. The left subfigure shows the similarity loss for both HERO and HDA, indicating that HERO-generated adversarial samples deviate minimally from the original, while the right subfigure displays prediction losses. Here, HERO outperforms HDA by identifying higher prediction errors with less disturbance introduced, thereby demonstrating superior effectiveness in generating critical test samples.

The graphical representation in Fig.~\ref{fig:compare_hda} confirms HERO's ability to maintain greater fidelity to the original sample and induce significant predictive errors. This analysis reinforces HERO's role in pinpointing and exploiting latent vulnerabilities within predictive models used in ICPS, further establishing its superiority in enhancing model robustness.
\subsection{Case 4 (advantages: Complexity perspective)}
For comparing complexiy perspective advantages of our proposed methd, we conducted a detailed computational complexity analysis for both the HERO using the ARO algorithm and the state-of-the-art HDA approach documented by Huang et al. \cite{huang2023hierarchical}. This analysis is pivotal in understanding the computational demands and efficiencies of both methods. The computational complexities of the usual GA can be mathematically modeled as follows, providing insight into the operational efficiency:
\begin{equation} \begin{aligned} O(GA) &= O(\text{initialization}) + O(\text{fitness evaluation}) \\ &\quad + O(\text{selection})  + O(\text{crossover}) + O(\text{mutation}) \\ &= O(n + Gn + Gn + Gnm + Gnm) \\ &= O(2Gnm + 2Gn + n) \end{aligned} \end{equation}
Transitioning from the GA's computational complexity, we next examine the computational demands of the ARO algorithm utilized in HERO:
\begin{equation} \begin{aligned} O(ARO) &= O(\text{initialization}) + O(\text{fitness evaluation}) \\ &\quad + O(\text{position updating in exploration}) \\&\quad + O(\text{position updating in exploitation}) \\ &= O(n + Gn + \frac{1}{2} Gnm + \frac{1}{2} Gnm) \\ &= O(Gnm + Gn + n) \end{aligned} \end{equation}
In these formulas, $n$ represents the number of individuals in the population, $G$ is the maximum number of iterations, and $m$ is the variable dimensionality of problems. These expressions succinctly encapsulate the computational dynamics of both algorithms, illustrating how HERO, powered by ARO, achieves superior time efficiency through strategic computation reduction.

\begin{figure}[ht]
    \centering
    \includegraphics[width=0.49\textwidth]{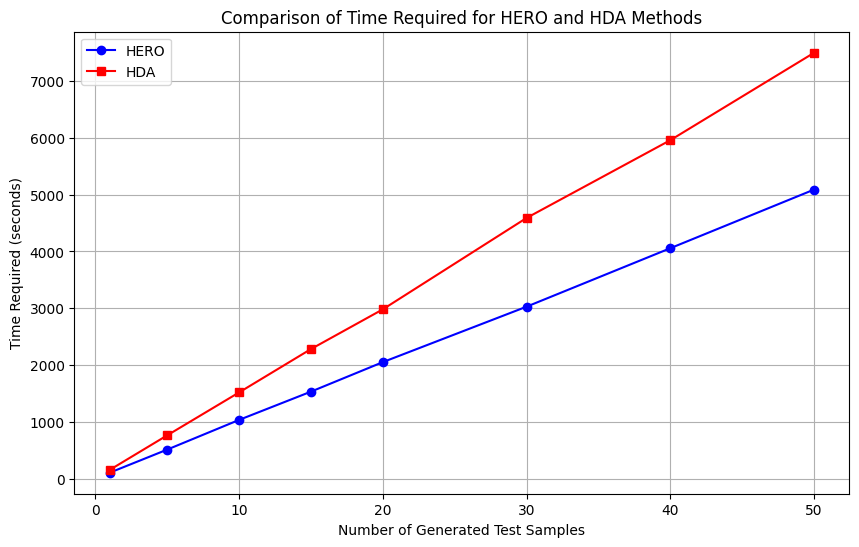}
    \caption{Comparison of Time Required for HERO and HDA Methods to Generate Test Samples under Same Conditions ($G$ = 100 and initial $N_{pop}$ = 300).}
    \label{fig:time_comparison}
\end{figure}

Following the complexity analysis, the experimental results in Fig.~\ref{fig:time_comparison} reveal that HERO consistently outperforms HDA in computational speed under identical experimental conditions, with the efficiency gap widening as the demand for test samples increases. This improvement is attributed to the ARO algorithm's strategic use of exploration and exploitation tactics, such as detour foraging and random hiding. These strategies minimize unnecessary evaluations and significantly reduce computational overhead, making HERO particularly advantageous in resource-constrained environments, offering more effective test case generation with reduced time consumption. While recognizing the time-efficiency enhancements brought about by the HDA algorithm over traditional coverage-guided testing methods, our findings highlight HERO's capacity to surpass HDA in time efficiency, providing more streamlined and potent adversarial testing. This advancement makes HERO an efficient tool for enhancing the reliability and security of ICPS.
\section{Discussion for Parameter Selection}
The selection of parameters within the HERO framework is guided by both theoretical foundations and practical requirements of industrial applications. The norm ball parameter for perturbation magnitude is constrained by physical system limitations to ensure adversarial examples remain within realistic operational boundaries. The ARO optimization parameters, including population size and iteration count, balance exploration thoroughness with computational efficiency to achieve optimal results within reasonable time constraints. The weighting factor between similarity and predictive loss ensures generated examples are both effective at exposing vulnerabilities and realistic enough to represent genuine operational scenarios. The number of test seeds is selected based on desired input space coverage and available testing resources. These parameters are determined through an iterative process involving theoretical justification and empirical validation, allowing HERO to be adapted across different industrial scenarios while maintaining optimal performance.

\section{Limitations and Future Work}
The efficacy of the HERO method hinges on assumptions about the input data distribution used to generate adversarial examples, emphasizing both global and local data characteristics. Significant discrepancies between these theoretical distributions and actual real-world data can compromise the realism of adversarial examples and the validity of our tests. Additionally, the inherent complexities and temporal dependencies within time series data require sophisticated analytical methods to accurately model and generate adversarial perturbations. This process is particularly challenging in real-time applications where computational efficiency is paramount, necessitating a balance between robust analysis and the constraints of low-latency environments.

Moreover, HERO's ability to produce physically plausible adversarial examples is limited by the accuracy of the physical constraints modeled within the system. These constraints rely on domain-specific knowledge, which may not fully capture the variability and complexity of operational conditions in intricate industrial settings. Moving forward, future work can explore adaptive and more computationally efficient algorithms that can handle large-scale data and refine our approach to better mirror the diverse and dynamic nature of real-world scenarios, enhancing the robustness and applicability across various implementation contexts.

\section{Conclusion}\label{sec:6}
In this study, we proposed Hierarchical Testing with Rabbit Optimization, a black-box adversarial testing algorithm for detecting the vulnerability of PHM systems in ICPS. HERO leverages ARO to generate physically constrained adversarial examples, combining global and local robustness perspectives to identify the most impactful samples that degrade model performance. The TS Transformer and PEMFC were chosen as the test cases due to the Transformer’s superior performance in normal conditions and multi-task adaptability, alongside the PEMFC’s dynamic operational variability and significance as a sustainable energy solution. Results reveal that despite its strong baseline performance, the TS Transformer remains highly vulnerable in adversarial scenarios, highlighting the urgent need for robust PHM systems. HERO’s generalizable framework extends its applicability to diverse ICPS domains, providing a foundation for advancing resilient PHM systems.

\end{document}